\documentclass[letterpaper, 10 pt, journal, twoside]{ieeeconf}  

\IEEEoverridecommandlockouts                              

\usepackage{amsmath} 
\usepackage{amssymb}  
\usepackage{xcolor}
\usepackage{array}
\usepackage{siunitx}
\usepackage{graphicx}
\usepackage{bm}
\usepackage{booktabs}
\usepackage{hyperref}
\usepackage[sorting=none,backend=biber]{biblatex}
\usepackage{color}
\usepackage{leftidx}
\usepackage{mathtools}
\usepackage{url}
\usepackage{changes}
\usepackage{makecell}
\usepackage{siunitx}
\title{
Lidar with Velocity: Correcting Moving Objects Point Cloud Distortion from Oscillating Scanning Lidars by Fusion with Camera
}

\author{Wen Yang$^{1}$, Zheng Gong$^{1}$, Baifu Huang$^{1}$ and Xiaoping Hong$^{1}$
\thanks{$^{1}$All authors are with the School of System Design and Intelligent
        Manufacturing, Southern University of Science and Technology, China. Zheng Gong is also with the School of Computer Engineering, Jimei University, China \{{\tt\small yangw2020@mail.sustech.edu.cn, kentz1988@gmail.com, huangbf@mail.sustech.edu.
        cn, hongxp@sustech.edu.cn\}} Correspondance to XH.}
}

\begin{document}

\maketitle

\begin{abstract}

Lidar point cloud distortion from moving object is an important problem in autonomous driving, and recently becomes more demanding with the emerging of oscillating type lidars, which feature back-and-forth scanning patterns and complex distortions. Accurately correcting the point cloud distortion would not only describe the 3D moving objects more accurately, but also enable accurate estimation of moving objects' velocities with enhanced prediction and tracking capabilities. A lidar and camera fusion approach is proposed to correct the oscillating lidar distortions with full velocity estimation. Lidar measures the time-of-flight distance accurately in the radial direction but only with sparse angular information while camera as a complementary sensor could provide a dense angular resolution. In addition, the proposed framework utilizes a probabilistic Kalman-filter approach to combine the estimated velocities and track the moving objects with their real-time velocities and correct point clouds. The proposed framework is evaluated on real road data and consistently outperforms other methods. The complete framework is open-sourced\footnote{https://github.com/ISEE-Technology/lidar-with-velocity} to accelerate the adoption of the emerging lidars.

\end{abstract}


\section{INTRODUCTION}

\begin{figure}[t]
  \centering
  \includegraphics[scale=0.18]{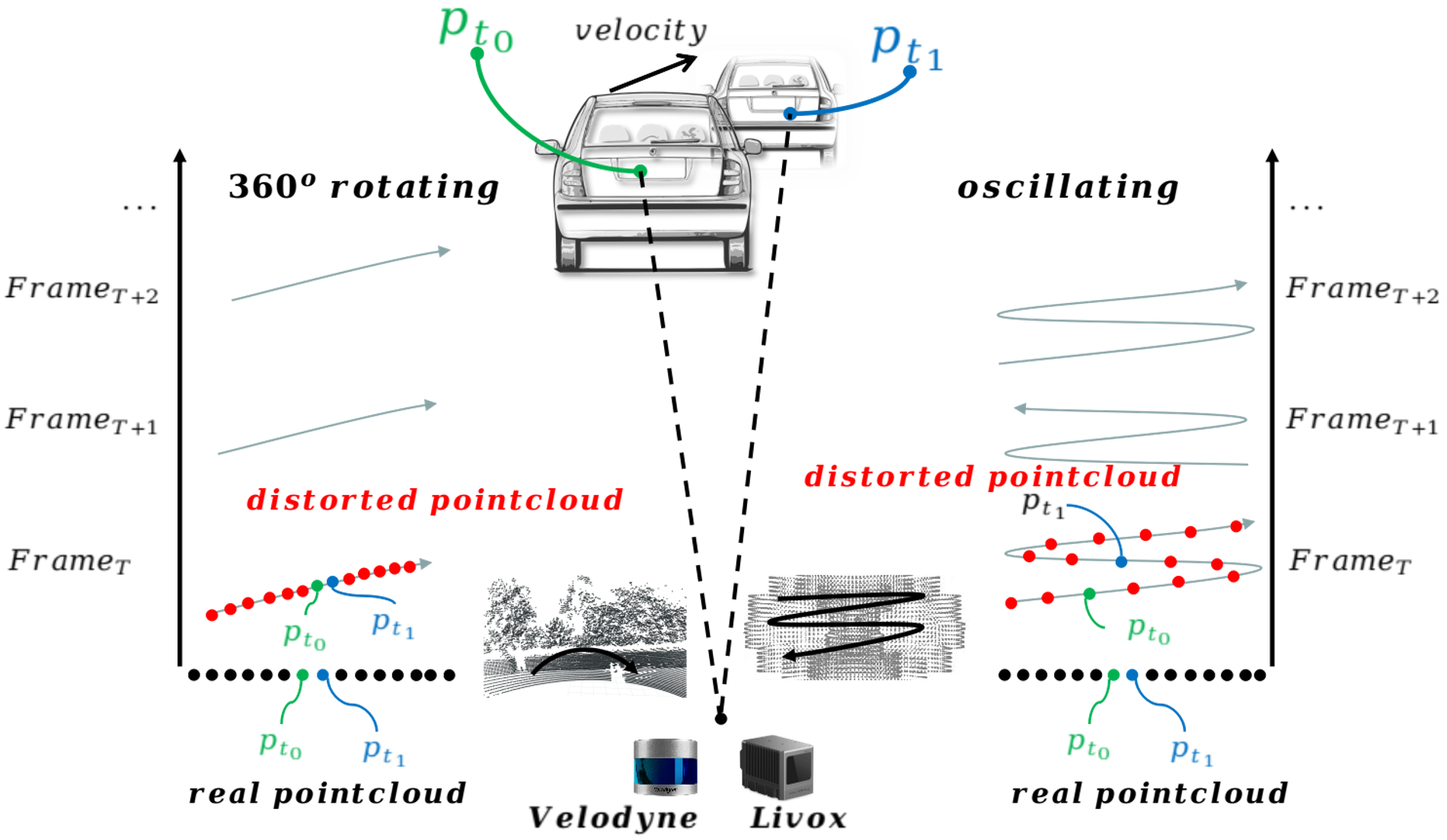}
  \caption{Typical distortion from \ang{360} rotation scan (Velodyne, left) and an oscillating scan (Livox Horizon, right), where oscillating scanning combined point clouds from various time and position into a single frame and degraded the motion distortion.}
  \label{fig:scanningPattern}
\end{figure}

\PARstart{A}{utomotive} lidar is playing an increasingly important role in modern autonomous driving as it provides direct and accurate 3D description of objects in the field of view (FoV). However the point cloud blurring typically happens when the objects are in motion. Similar to the mechanism of motion blur in rolling-shutter cameras, the distortion stems from the lidars' scanning nature as seen in Fig. \ref{fig:scanningPattern}; progressive time-of-flight (ToF) measurements take place one after another while the observed object moves during these measurements. Correcting these distortions does not only give a better 3D description of the object, but also serves as a means to evaluate accurately the velocity of the moving object at each instance. Full velocity information (e.g. a 4D lidar) and the associated corrected point cloud distortion are highly desired to help in object recognition, pose identification, tracking, motion prediction and decision making \cite{azim2012detection, held2013precision}. The distortion from the traditional \ang{360} rotating lidar has been observed in previous works \cite{hong2010vicp,feldman2012the,levinson2011towards}, where the phenomenon is usually a deformation of the object (left part of Fig. \ref{fig:scanningPattern}). However due to the higher performance, reliability and cost requirements from the automotive industry, emerging lidars are developed \cite{liu2021low} and possess more severe motion distortions (Table \ref{table:table_vendors}), where the patterns feature back-and-forth angular scanning in a single frame (i.e. oscillating scan) and the resulted blurring is enlarged through the extended duration and scale of the ToF measurements, as shown on the right part of Fig. \ref{fig:scanningPattern}.

\begin{table}[htb]
\centering
\caption{Different types of lidars, representative manufacturers, known adopting automotive OEMs and distortion severity.}
\label{table:table_vendors}
\begin{tabular}{ |m{2.5cm}|m{1.6cm}|m{1.2cm}|m{1.2cm}|}  \hline
Lidar Type & Representative & Adopting OEM & Distortion Severity \\ [0.5ex] 
 \hline 
 Flash & Continental &  & None/Mild \\ 
 \hline 
 \ang{360} Rotation & Velodyne &  &Moderate\\ 
 \hline 
 Micro Motion & Cepton & GM  & Severe \\
 \hline 
 Scan/Rotating Mirror & Luminar & Volvo & Severe \\
 \hline 
 MEMS Mirror & Innoviz & BMW & Severe \\
 \hline 
 Rotating Prism & Livox & Xpeng & Severe \\ 
 \hline
\end{tabular}
\end{table}

\begin{figure}[htbp]
  \centering
  \includegraphics[scale=0.17]{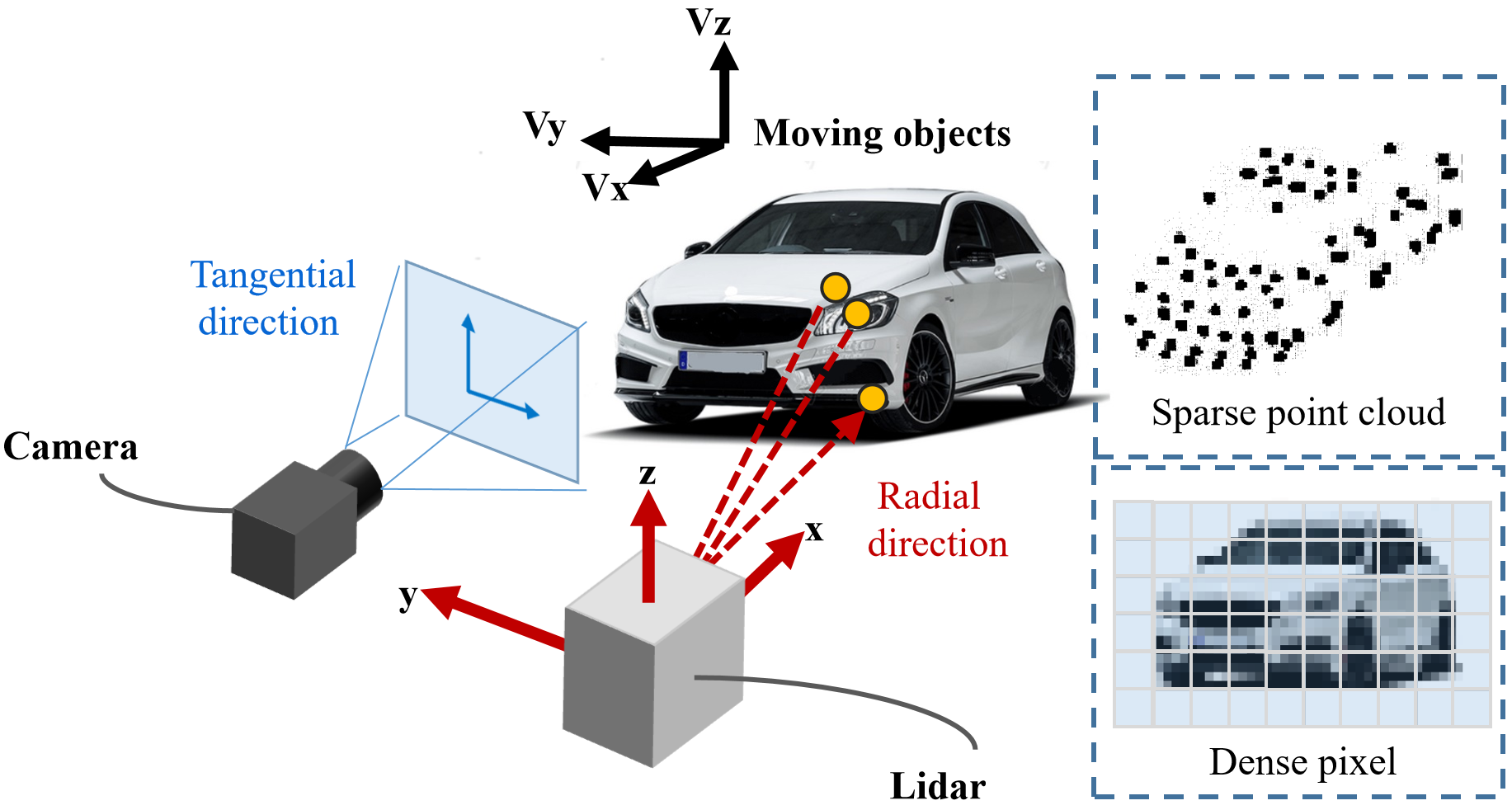}
  \caption{Lidar has a good precision in radial direction, but lacks in angular resolution. Camera has a good angular resolution, but cannot directly detect distance. Fusion of them could provide complementary measurements on the 3D object.}
    \label{fig:lidarCamera}
\end{figure}

Full velocity estimation is not trivial using lidar points alone. In principle the velocity can be extracted by aligning point clouds from successive frames. However due to limited lidar angular resolution, point cloud resolution is lacking in the tangential direction (i.e. polar and azimuthal directions in spherical coordinate as in Fig. \ref{fig:lidarCamera}) and it is difficult to align the point cloud tangentially. The additional blurring caused by the oscillating scan exaggerates its difficulty. On the other hand, cameras provide essentially high angular resolution measurements, but no direct distance (radial) measurement (Fig. \ref{fig:lidarCamera}). For example, a standard 1080p camera with 2 million pixels per frame easily outperforms the state-of-the-art Velodyne HDL-64E lidar with 0.13 million points per frame. Therefore, fusing camera and lidar together for a full 3D velocity estimation combines the strengths of both. In this paper, we devise a sensor fusion approach to simultaneously identify the full velocity of the moving objects as well as recover the blurred point clouds from oscillating lidars, with the following contributions.

\begin{enumerate}
  \item This is the first work correcting the moving object distortions from the emerging oscillating type lidars. A probabilistic sensor fusion algorithm combining the respective strengths of camera and lidar is proposed to correct the moving objects' point cloud distortion.
  \item A full and accurate 3D velocity is estimated and combined in a Kalman filter system, enabling accurate moving object prediction and tracking.
  \item A complete framework solution from the frontend sensor detection to the backend tracking with real time performance is provided and open-sourced to accelerate the adoption and development of the emerging lidars. 
\end{enumerate}

\section{RELATED WORK}

Lidar point cloud distortion is usually separated into its ego-motion distortion and object motion distortion. The ego-motion distortion can be simply corrected by an IMU. However due to the noise and drift issues from the low-cost IMUs, advanced methods such as simultaneous localization and mapping (SLAM) have been used to provide a much more robust ego-motion information. For example, Zhang \textit{et al.} \cite{2014LOAM} assumed a uniform motion relative to still objects to correct the self-motion distortion within the lidar frame, and achieved good performance in both mapping and localization accuracy.  Although the ego-motion based mechanism can effectively correct the point clouds of the still scenes, it does not correct the moving objects. Velocity estimations of these moving objects are indispensable in distortion correction, and different methods have been proposed to solve this problem. From the perspective of sensor types, these methods are roughly divided into camera based, lidar based, and sensor-fusion based methods.

With cameras, a few different approaches are widely used to estimate the velocity of moving objects from images. The optical flow \cite{1991Carlo,1997An} approaches including Lucas-Kanade and KLT assume the flow is continuous in a local neighbourhood and the optical flow equations for all the pixels in that neighborhood were used to solve for a 2D velocity. Direct learning based methods have also been proposed to estimate the optical flow such as FlowNet\cite{2017FlowNet}, PWCNet\cite{2017PWC} and EPIFlow\cite{2008Epiflow}. However the optical flow does not provide a true scale and lacks in 3D velocity estimation. Detection based methods can be used to identify the object type with a region of interest (ROI) \cite{2018YOLOv3,2020YOLOv4} and give a hint of the object scale. End-to-end methods such as \cite{kampelmuhler2018camera} directly estimate vehicle velocities from their trajectories using a multilayer perceptron. However since the depth information is not directly obtained, object depth and the associated velocity might not be accurate.

With lidars, point cloud tracking methods could provide 3D motion of the object. Petrovskaya \textit{et al.} \cite{petrovskaya2009model} extracted anchor points to label the objects and used these features for tracking. Baum \textit{et al.} \cite{baum2014extended} used geometric shape fitting such as ellipse to simplify the point cloud feature description. More general methods aligning the point clouds such as ICP are also widely used, but the optimizations are prone to initial condition and local minima \cite{held2013precision}. 

Lidar and camera provide respectively excellent depth information and angular resolution. Based on the complementary characteristics, sensor fusion is evidently advantageous for object detection, tracking and velocity estimation. Held \textit{et al.} \cite{held2014combining} used annealed dynamic histograms to globally explore the state space for object velocity estimation, and an extended model was used to integrate RGB information to track moving objects. The distortion correction from this method also achieved excellent performance. Zhang \textit{et al.} \cite{0Vehicle} extracted the ROI and estimated the objects' contour parameters by utilizing the random hypersurface models (RHM) through camera and lidar fusion. Daraei \textit{et al.} \cite{daraei2017velocity} proposed a tightly-coupled fusion approach by considering several energy terms related to camera and lidar respective characteristics in the optimization. 

Our objective aligns with these previous efforts in trying to accommodate the point cloud distortion arising from the motion of the object. Most of the previous work considers the distortion arising from successive frames (inter-frame) while ignoring the distortions occurred within each frame (intra-frame), as they are relatively small for the \ang{360} rotating lidars. Assuming a moving object has \ang{20} angular extension and lidar operates at 10 Hz (100 ms frame duration), the total scan time of the object in one frame is roughly $100 {\:\rm ms} \times (20\si{\degree}/360\si{\degree}) = 5.6 {\:\rm ms}$. On the other hand, the oscillating lidars could have points describing that object coming from anywhere of the 100 ms frame duration (Fig. \ref{fig:scanningPattern}). As a result, the previous methods did not accommodate these severe intra-frame distortions from the oscillating type lidars. Our approach recognizes distortion from each lidar point and corrects it by minimizing the spatial distribution of the moving object's total point cloud. Correcting the oscillating scanning type distortion has not been explored previously but becomes an important problem in emerging automotive lidar adoption.
\section{SYSTEM AND METHOD}

\subsection{Overview}\label{Overview:platform}
\begin{figure}[htbp]
  \includegraphics[scale=0.30]{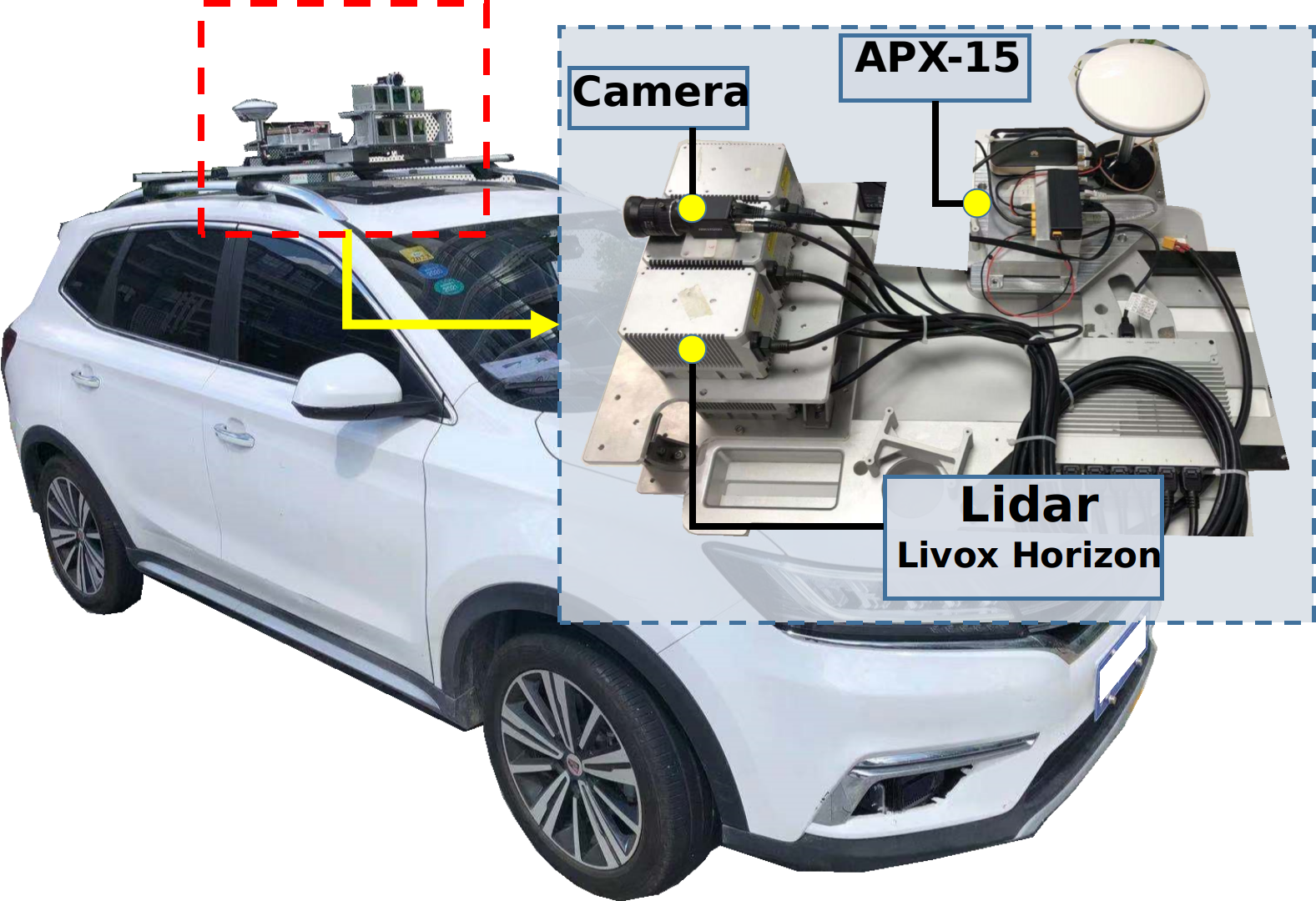}
  \caption{The hardware system is composed of Livox Horizon lidars, a RGB camera and a GNSS-Inertial module. This system is mounted on a moving vehicle to gather data from real road scenarios.}
  \label{fig:system}
\end{figure}

Our hardware system is illustrated in Fig. \ref{fig:system}. We choose Livox Horizon lidars \cite{liu2021low} as an example of oscillating scanning lidars, with scanning pattern similar to the lower-right pattern in Fig. \ref{fig:scanningPattern}. The Livox Horizon lidars were mounted on top of the moving vehicle. A RGB camera is mounted with the same FoV as lidars to detect the moving objects and estimate their tangential movement velocity. A GNSS-Inertial system (APX-15) is utilized to accurately measure the ego-motion for the corresponding distortion correction. The lidars, the camera and the APX-15 module are synchronized in time. 

\begin{figure} [htbp]
  \includegraphics[scale=0.42]{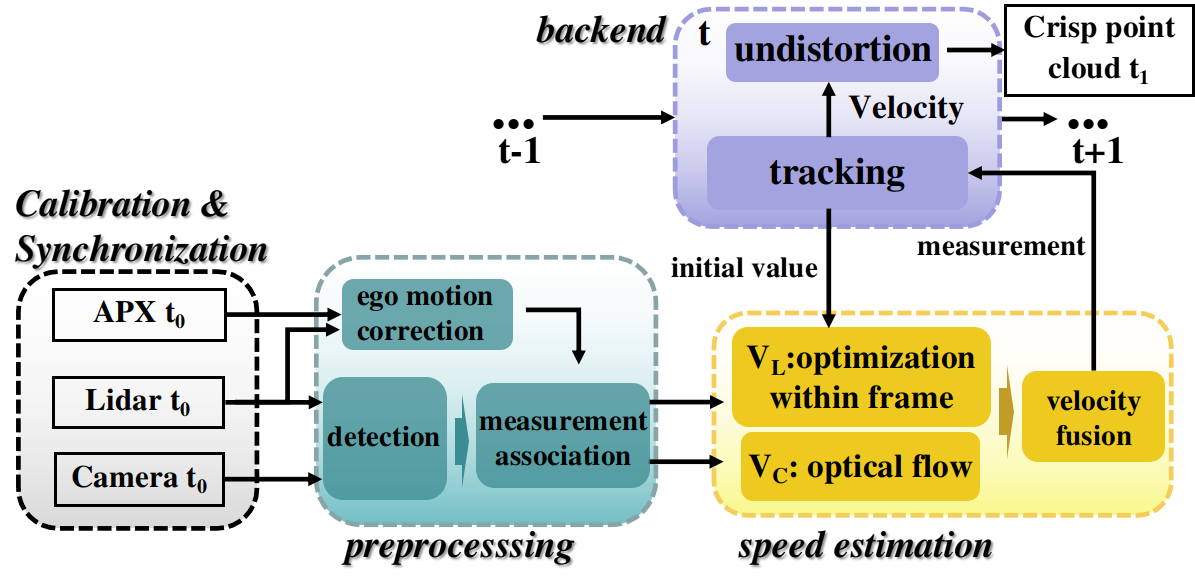}
  \caption{The pipeline of the framework combining the strengths of both lidar and camerea with a Kalman filter based tracking.}
  \label{fig:pipeline}
\end{figure}

With this set of sensor hardware, we proposed a systematic framework to estimate the moving objects velocities and correct their point cloud distortions, with a systematic pipeline as illustrated in Fig. \ref{fig:pipeline}. The extrinsically calibrated \cite{zhu2020camvox,yuan2021pixel} lidar and camera fed sensor data into the preprocessing stage, which contains ego-motion correction, object detection and association. The ego-motion correction is performed using either the APX inertial system or the lidar odometry, both of which have been provided in the framework. In detection, the moving objects are identified with YOLO (a deep-learning CNN-based object detection algorithm) \cite{2020YOLOv4} or Livox Detection (a deep-learning point cloud detection algorithm) \cite{livoxdetecion} or both. 

Once an object from the image is identified, the corresponding point cloud is extracted and associated. Separate optimizations are performed on both the point cloud data and the image data for 3D velocity estimation and tangential velocity estimation respectively at frame update rate. These two velocities are fused probabilistically into a unified velocity, which is fed into a Kalman filter (KF) process as a measurement. The KF process runs for each tracked object to compute its final velocity. With this velocity the distorted point cloud of each moving object is corrected. The point cloud quality is evaluated with a crispness score if necessary.

\subsection{Pre-processing}

We correct the moving object distortion in the global coordinate, where the distortions from the ego-motions (rotation and translation) are removed. Calibrations are performed for all the sensors to transform them into the global coordinate. They include the lidar camera extrinsic and the lidar-APX extrinsic calibrations. Time synchronization is also performed at the hardware level. Since all the sensors are synchronized, for ego-motion we can correct the point cloud data gathered during a single camera frame (usually 100 ms) to the start of the frame at ${t_0}$. The APX poses (rotation, translation) at the start of current frame $(\textbf{R}_{t_0}, \textbf{T}_{t_0})$ and next frame $(\textbf{R}_{t_1}, \textbf{T}_{t_1})$ are used to transform each lidar point according to a constant velocity motion model assumption of our vehicle in a short time.

The ego-motion undistorted point $\textbf{P}_i$ for the $i$th point with timestamp $t_i$ in global coordinate can be formulated from raw lidar point in its own coordinate $\textbf{P}^{'}_{i}$ as:
\begin{equation}
	\textbf{P}_i = \textbf{R}_{t_i}\textbf{P}^{'}_{i} + \textbf{T}_{t_i}
	\label{formular:undistortion_by_velocity}
\end{equation}
where $\textbf{R}_{t_i} = \textbf{R}_{t_0}\textbf{R}_{t_0,t_i}$ and $\textbf{T}_{t_i} = \textbf{R}_{t_0}\textbf{T}_{t_0,t_i} + \textbf{T}_{t_0}$ represent the global rotation and translation of sensor suite at time $t_i$. Here the relative rotation and translation between $t_i$ and $t_0$ are obtained from interpolation:
\begin{equation}
	\textbf{R}_{t_0,t_i}=e^{\hat{\omega}\theta\frac{t_i-t_0}{t_1-t_0}},   \textbf{T}_{t_0,t_i}=\frac{t_i-t_0}{t_1-t_0}\textbf{T}_{t_0,t_1}
\end{equation}
where $\theta$ represents the magnitude of the rotation and $\hat{\omega}$ is the unit vector in rotation direction. The rotation interpolation can be derived by Rodrigues formula  \cite{Murray1994AMI}:
\begin{equation}
\begin{aligned}
	\textbf{R}_{t_0,t_i} = 
	e^{\hat{\omega}\theta} = 
	\textbf{I} + \hat{\omega}\sin{\theta} + {\hat{\omega}}^2(1-\cos{\theta})
\end{aligned}
\end{equation}

Alternatively if an accurate GNSS-Inertial system is not available, a lidar-inertial-odometry system can be utilized to estimate ego-motion velocity. 

After removing the ego-motion distortion, the moving objects and a corresponding ROI bounding box are detected through the image by the YOLO algorithm \cite{2020YOLOv4}. The corresponding point cloud of the object is segmented through the bounding box and associated with the object. If the camera detection is not available, one could also employ a 3D point cloud detection algorithm \cite{livoxdetecion} which we have also provided in the framework.

\subsection{Tangential Velocity Measurement From Camera}\label{sec:Tangential_Velocity_Measurement_From_Camera}

\begin{figure}[htbp]
  \includegraphics[scale=0.18]{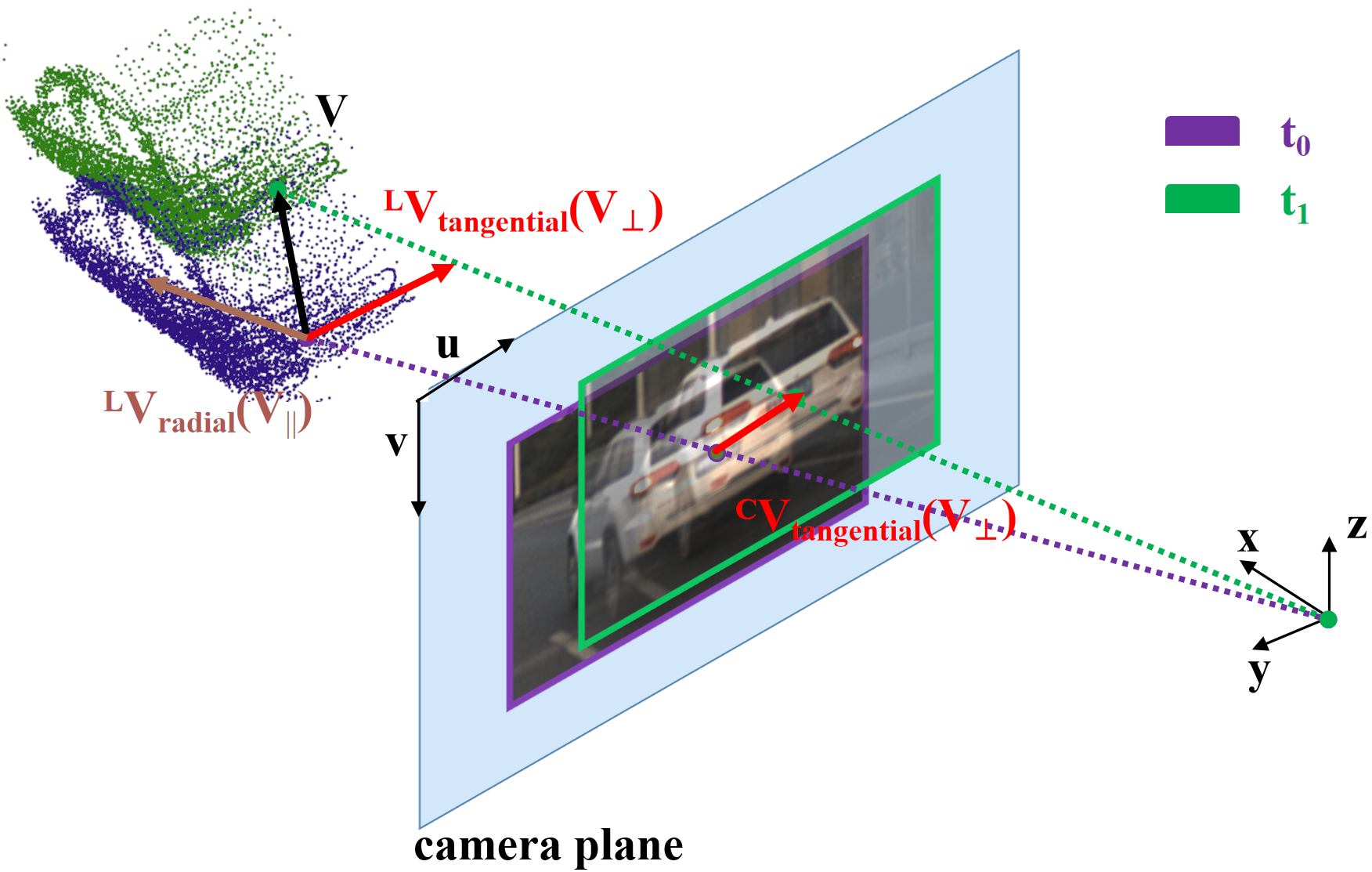}
  \caption{The velocity measurement can be decomposed into two orthogonal directions. Cameras can measure tangential (polar and azimuthal directions in spherical coordinate) velocities more accurately due to higher resolution, while lidars could measure both tangential and radial directions. Fusion of two sensor modalities in tangential directions is beneficial.}
  \label{fig:velocityfusion}
\end{figure}

The camera provides a good tangential velocity measurement mechanism. As illustrated in Fig. \ref{fig:lidarCamera}, clearly the image resolution is much higher to determine the tangential velocity accurately than using the point cloud where the angular resolution is sparse. For two consecutive frames at time $t_0$ and $t_1$, the motion is relatively small and a KLT sparse optical flow algorithm \cite{1997An} is performed to track the pixel motion field. Since the detection bounding box for each moving object is available, the optical vectors within the bounding box indicate the relative 2D movement of the object. The RANSAC method \cite{hartley2000multiple} is used to remove the outliers and refine the results. The distribution of the pixel velocity vectors can be formulated as a 2D Gaussian distribution ${{v}_{c}^{'}}\sim \mathcal{N}(\textbf{v}_{c}^{'}, \mathbf{\Sigma}_{c}^{'})$ where the 2D mean and covariance matrix is evaluated from the optical flow velocity vector set in the camera coordinate. Following the computation by Daraei \textit{et al.} \cite{daraei2017velocity}, the projection from 2D velocity 
$\textbf{v}_{c}^{'} = (v_\theta,v_\varphi)^T$ to 3D velocity $\textbf{v}_{flow}^{'}$ is computed in the camera relative coordinate. 

\begin{equation}
\begin{split}
	{\textbf{v}_{flow}^{'}({\textbf{v}_{c}^{'}})} = d \left [ 
    \begin{array}{cc}
        1/f_\theta & 0 \\
        0 & 1/f_\varphi \\
        0 & 0
    \end{array} 
    \right ]
    \left [ 
    \begin{array}{c}
    v_\theta \\
    v_\varphi
    \end{array}
    \right ]
\end{split}
\end{equation}

Here $d$ represents the object depth averaged by the object point clouds, $f_\theta$ and $f_\varphi$ are the camera focal lengths along the two tangential directions.  

The relative velocity is then transformed to 3D global coordinate $\textbf{v}_c$ by differentiating Eqn.  \ref{formular:undistortion_by_velocity}. The resulting equation comprises of two terms. One is related to the relative velocity $\textbf{v}_{flow}^{'}$ computed above, and the other is related to the sensor ego-motion, considering sensor velocity and angular velocity. 

\begin{equation}
\begin{split}
	{\textbf{v}_c}  = & \mathcal{B}({\textbf{v}_{c}^{'}}) 
	\\ = & \textbf{R}_{t_0}\textbf{v}_{flow}^{'}({\textbf{v}_{c_0}}) + \textbf{v}_{ego}(d)
\end{split}
\end{equation}

It is noted that although a 3D velocity in the global coordinate is computed, the information from image is limited to the tangential velocity (polar and azimuthal directions) as shown in Fig. \ref{fig:velocityfusion}. The relative radial velocity is not estimated. In Section. \ref{sec:fusion}, we will project this $\textbf{v}_{c}$ onto the tangential direction anyway.

Since the term $\textbf{v}_{ego}$ is independent on target object velocity, the covariance of the 3D velocity distribution ${v}_c\sim \mathcal{N}(\textbf{v}_c,\mathbf{\Sigma}_c)$ can be formulated as follow:
\begin{equation}
\begin{aligned}
	 \mathbf{\Sigma}_c & = 
	\textbf{J}_{\mathcal{B}}\bm{\Sigma}_{c}^{'}\textbf{J}_{\mathcal{B}}^T
\end{aligned}
\end{equation}

where $\textbf{J}_{\mathcal{B}}$ is the Jacobian matrix of function $\mathcal{B}$, and $\mathbf{\Sigma}_{c}^{'}$ indicates the covariance matrix of the optical flow velocity vector set in the camera coordinate.

\subsection{Lidar Measurement and Optimization}\label{Lidar_Measurement_and_Optimization}
Lidar provides another modality from the camera. The contribution of the lidar are two folds. First, the direct distance measurement allows one to estimate the movement of the object along the radial (depth) direction. This is missing from the camera modality. Second, although relatively sparse, the lidar points could still be used to estimate the tangential velocity (Fig. \ref{fig:velocityfusion}), with computed covariances that is ready to be fused with camera modality. We propose a novel optimization method to estimate object full velocity from lidar. This velocity will be projected onto the radial direction and the tangential directions. The radial part will be used as the estimated object radial velocity while the tangential part will be fused with the camera measurement to combine strength from both modalities.

Our optimization is especially suited for the oscillating pattern of the new lidars with severe motion distortion (e.g. Livox Horizon) where we take account the deviation of each point. The back-and-forth scan within one frame leads to blurred multi-layer point cloud contour in moving object. Since each point is obtained at a precise and distinct timestamp, the correction can be performed with an accurate velocity estimation to form a clear point cloud, which has a narrower distribution in 3D space than original. The above holds true for both local scale and global scale of the entire object. The point cloud associated with the object is firstly divided into small voxels of given sizes. The variance of the point cloud is supposedly minimized within each voxel. The global variance is also minimized to suppress multiple shadows of the same object. Omitting the rotation of the object with respect to the sensor, the minimization problem is formulated as following:

\begin{equation}
\begin{aligned}
	\mathop{\arg\min}\limits_{\textbf{v}}
	\bigg\{
	\overbrace{
	{\sum\limits^{k\in {\textbf{G}}}_k} \ 
	E_k
	}^{local}
	+ 
	\overbrace{
	E_g
	}^{global}
	\bigg\}
\end{aligned}
\end{equation}

\begin{equation}
	E_k, E_g = 
	{\sum\limits_{i}}
	{\sum\limits_{r}}
	\left \|
	\left(
	(\textbf{P}_{i}-\Delta t{\textbf{v}}) - \overline{\textbf{P}}
	\right)
	\cdot \textbf{n}_{\textbf{r}}
	\right \|^2
\end{equation}

Here $\textbf{G}$ is the voxel grid set constructed from the input point cloud. The $r$ represents the three projection orthogonal basis. $\textbf{n}_{\textbf{r}}$ denotes the unit vector in the \textbf{r} direction of three orthogonal directions. $E_k, E_g$ represents the variance of points in the voxel set or the global set. $\overline{\textbf{P}}$ is the undistorted point cloud centroid in that corresponding set. $\Delta t$ represents the difference between the collection time of point $\textbf{P}_{i}$ to the frame start time. Ceres Solver \cite{ceres-solver} is used in optimization to find out the optimal ${v} \sim \mathcal{N}(\textbf{v},\bm{\Sigma})$. 
The covariance of the optimal velocity $\textbf{v}$ can be derived as $\bm{\Sigma_l} = J'({\textbf{v}}){S^{-1}}J({\textbf{v}})$. $S$ denotes the covariance of the observed points in the set. $J$ is the Jacobian of cost function at $\textbf{v}$. 

\subsection{State Fusion and Tracking} \label{sec:fusion}

In each frame the 3D velocity $\textbf{v}_f$ is estimated from the two modalities, ${v}_l  \sim \mathcal{N}(\textbf{v}_l,\bm{\Sigma}_l)$ from lidar, and ${v}_c \sim \mathcal{N}(\textbf{v}_c,\bm{\Sigma}_c)$ 
from camera. Note $\textbf{v}_c$ has information only in tangential directions. The final fused velocity $v_f$ served as a measurement input to the Kalman filter at frame output rate.

The radial and tangential components are firstly separated. The velocity distribution for 
${v}^{\parallel}_l\sim(\textbf{v}^{\parallel}_l,{\bm{\Sigma}_l^{\parallel}})$, ${v}^{\perp}_l\sim(\textbf{v}^{\perp}_l,
{\bm{\Sigma}_l^{\perp}})$ and  ${v}^{\perp}_c\sim(\textbf{v}^{\perp}_c,
{\bm{\Sigma}_c^{\perp}})$ are obtained by projecting onto the radial and tangential direction respectively, with formula:

\begin{equation}
    \textbf{v}_{proj} = h(\textbf{v}), {\bm{\Sigma}_{proj}}=\textbf{H} {\bm{\Sigma}} \textbf{H}^{T}
\label{formula:projection}
\end{equation} 
where $h$ is $h^{\parallel}$ or $h^{\perp}$, $H$ is the Jacobian matrix of the projection function $h$ derived at velocity $\textbf{v}$. 

In radial direction, the velocity optimized with lidar is trusted. In tangential direction, the velocity estimated by camera and by lidar are fused according to their covariances. Therefore the fused velocity measurement $\textbf{v}_f$ in both directions can be derived as:

\begin{equation}
\begin{aligned}
    \textbf{v}^{\parallel}_{f} & = 
    \textbf{v}^{\parallel}_{l} \\
    \textbf{v}^{\perp}_{f} & = 
    \textbf{v}^{\perp}_{l} \boxplus 
    \textbf{v}^{\perp}_{c}
\end{aligned}
\end{equation}
here $\boxplus$ means fusing two Gaussians using their covariance:

\begin{equation}
\begin{aligned}
    \textbf{K} & =   {\bm{\Sigma}_l^{\perp}}
    ({\bm{\Sigma}_l^{\perp}}+{\bm{\Sigma}_c^{\perp}})^{-1} \\
    \textbf{v}^{\perp}_{f} & = 
    \textbf{v}^{\perp}_{l} + 
    \textbf{K}(\textbf{v}^{\perp}_{c} - \textbf{v}^{\perp}_{l})
\end{aligned}
\end{equation}
here $\textbf{K}$ represent the fusion gain between two independent Gaussians, similar to Kalman gain in Kalman filter. 

Finally $\textbf{v}_{f}$ is combined from $\textbf{v}^{\parallel}_{f}$ and $\textbf{v}^{\perp}_{f}$ as the input measurement value to the tracking backend process depicted in Fig. \ref{fig:pipeline}. 

Because in each step we have obtained the measurements and the associated uncertainty, it is natural to use Kalman filter for tracking due to its built-in uncertainty processing mechanism. Other filtering methods such as model-based ones might be better predicting the movements of each moving object species (car, pedestrians or bicycles) and could be included in further studies. We apply the Kalman filter based tracking mechanism similar to Weng \textit{et al.} \cite{weng2020ab3dmot} for state estimation and data association. The Kalman state space contains a 11-dimensional vector $\textbf{T}_{pre}=(x,y,z,\theta,l,w,h,s,v_x,v_y,v_z)$ including the object center$(x,y,z)$, heading angle $\theta$, object size $(l,w,h)$, the detection confidence score $s$ from YOLO and the velocity $(v_x,v_y,v_z)$. The data association between consecutive frames is evaluated by 3D IOU metrics and Hungarian algorithm \cite{weng2020ab3dmot}. The objects between the consecutive frames with the highest similarity are assigned with the same object ID.

\subsection{Undistortion Module for Moving Object Points}

Since the velocity of each moving object is predicted, the undistorted point for each moving point $\textbf{P}_{i}$ is corrected with the velocity to the frame start time $\textbf{t}_{0}$:
\begin{equation}
	\textbf{P}_{is} = 
	\textbf{I}\textbf{P}_i + (-\textbf{T}^{o}_{t_0,t_i})
	\label{formular:degenerate_undistortion}
\end{equation}
Here we assume that the moving object has movements only in translation but not in rotation. $\textbf{I}$ represents a identify rotation matrix. $\textbf{T}^o_{t_0,t_i}$ is the object translation from time $t_0$ to $t_i$ in global frame, which is expressed as $\textbf{T}^o_{t_0,t_i}=(t_i-t_0)\textbf{v}$. When the object is stationary, $\textbf{P}_{is}$ is the same as $\textbf{P}_i$.

\section{EXPERIMENTS}

\subsection{Camera for Tangential Resolution Enhancement}

\begin{figure}[htbp]
  \includegraphics[scale=0.22]{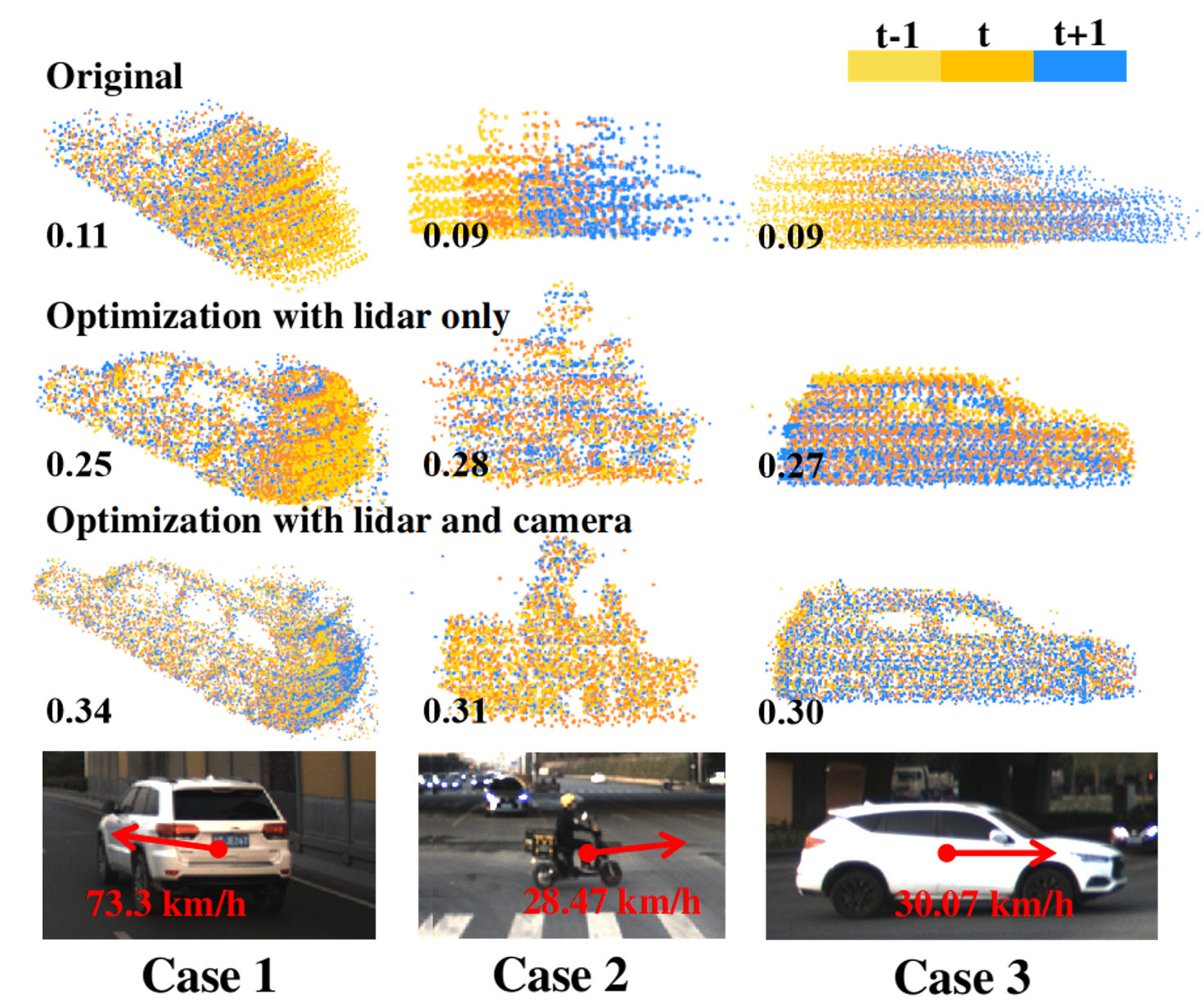}
  \caption{Point cloud distortion correction from three consecutive frames using the proposed method with only lidar point cloud optimization and lidar with camera optimization. The original point clouds are shown at the top and the corresponding images are displayed at the bottom. The numbers on the picture denote the respective crispness scores.}
    \label{fig:multi_frame_integration}
\end{figure}

Fig. \ref{fig:multi_frame_integration} demonstrates tangential resolution enhancement capability from cameras in point distortion correction. The original distorted point clouds from three consecutive frames are displayed on the first row. Due to the oscillating lidars' scanning nature, the blurring effect is significant. When optimizing with only point clouds from lidar, corrections are obviously lacking in the tangential direction, as compared to the case when optimizing from both lidar and camera, especially in cases where the moving object is moving mainly in the tangential direction. Kalman filter is used as the backend for all the cases to maintain a smooth velocity estimation. 

In order to quantitatively evaluate the point cloud distortion correction, a crispness score \cite{sheehan2012self} can be calculated to represent the quality of the undistorted point cloud, as defined below \cite{sheehan2012self}:

\begin{equation}
    \frac{1}{T^2}\sum_{i=1}^{T}\sum_{j=1}^{T}\frac{1}{n_i}\sum_{k=1}^{n_i}\bm{\mathop{G}}(\bm{\mathcal{P}}_k^i-\bm{\mathcal{P}}_k^j, \bm{\Sigma})
\end{equation}

where the T is number of the frames for which the object is detected and tracked, $n_i$ is the number of points in Frame $i$, $\bm{\mathcal{P}}_k^i$ is the point in Frame $i$ nearest to point $\bm{\mathcal{P}}_k^j$ in Frame $j$. $\bm{\mathop{G}}$ represent the multi-variate Gaussian distribution and $\bm{\Sigma}$ is the relaxation coefficient to control the extend of the boundary. Larger crispness value means better distortion correction and more accurate velocity estimation. The crispness score is evaluated and labeled for each case in Fig. \ref{fig:multi_frame_integration}.

\begin{figure}[htbp]
    \flushleft
  \includegraphics[scale=0.145]{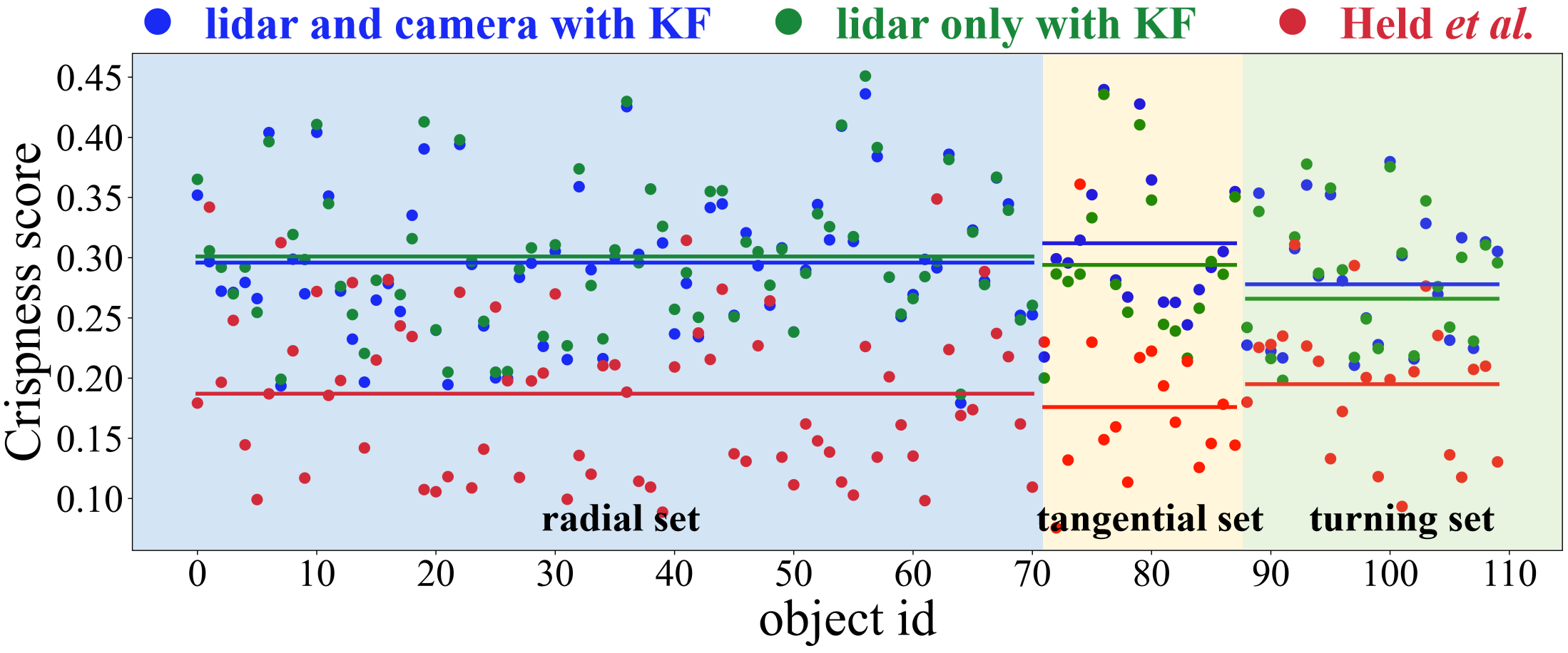}
  \caption{The crispness score of each object from the dataset consisting of 110 objects.}
    \label{fig:our_crisp}
\end{figure}

\begin{table}[htbp]
	\centering 
	\caption{average crispness score in different scenarios}
	\resizebox{0.50\textwidth}{!}
	{
  	\begin {tabular}{c c c c}

  		\toprule[1.0pt]	
  		-     &  {\textbf{\textit{lidar and camera with KF}}}  & {\textbf{\textit{lidar only with KF}}}   & {\textbf{\textit{Held}} \cite{held2014combining}}    \\ 
  		
  		\midrule[0.8pt]
  		radial set    & \makecell[c]{${0.296}$} & \makecell[c]{$0.301$}   & \makecell[c]{$0.187$}  \\ 
  		\midrule[0.3pt]
  		tangential set  & \makecell[c]{$0.312$}   & \makecell[c]{${0.294}$} & \makecell[c]{$0.176$}  \\ 
  		\midrule[0.3pt]
  		turning set   & \makecell[c]{$0.278$}   & \makecell[c]{${0.266}$} & \makecell[c]{$0.195$}  \\ 
  		\bottomrule[1.0pt]
  	\end{tabular}
  	}
\label{table:crispness_evaluation_subsets}
\end{table}

\subsection{Datasets Evaluation}\label{sec:Distortion_Correction_Evaluation}

40 sets real road data from different scenes were collected with the home-built vehicle-roof-mounted system (Fig. \ref{fig:system}). Each data set contains at least 15 seconds of driving scenario. There are 244 objects being tracked in this dataset with a total of 4732 tracked frames. The detection failed 14 frames out of the 4732 frames, but the tracking process could continue to work for single missed frames. The average 3D IOU for the tracking is 0.72 with a standard deviation of 0.1. Overall the tracking performs well with this framework.

To ensure the best evaluation on velocity estimation, 110 moving objects are selected from these object with the criteria that the observation time in the scene is larger than 2 seconds. These objects are then divided into radial (71 objects), tangential (17 objects), and turning (22 objects) subsets to evaluate undistortion performance in these different situations. The radial and vertical sets refers to the motion direction of the target objects, while the turning set refers to the sensor vehicle itself when it is rotating.

The method proposed by Held \textit{et al.} \cite{held2014combining} is a state-of-the-art widely adopted algorithm in correcting the lidar point cloud distortions from the traditional \ang{360} rotating type lidars. The proposed method (lidar and camera, lidar only) is evaluated against the Held \textit{et al.} method, and the result for each individual object is shown in the Fig. \ref{fig:our_crisp}. The average crispness scores from these three subsets are indicated as lines in the figure and also presented in Table \ref{table:crispness_evaluation_subsets}. Clearly the proposed methods outperform by a large margin compared to the Held \textit{et al} method, which did not consider the intra-frame motion distortion. Additionally, tangential subset and turning subset have better evaluations from the sensor-fusion approach than from the lidar-only approach, consistent with our expectations that the large camera resolution helps in the angular estimation. The turning subset overall performance is slightly lower than the other two subsets probably due to the omission of rotation in the object vehicle motion model and the errors introduced in estimation of ego-motion rotation.

\subsection{Ground Truth Comparison}

\begin{figure}[htbp]
    \flushleft
  \includegraphics[scale=0.32]{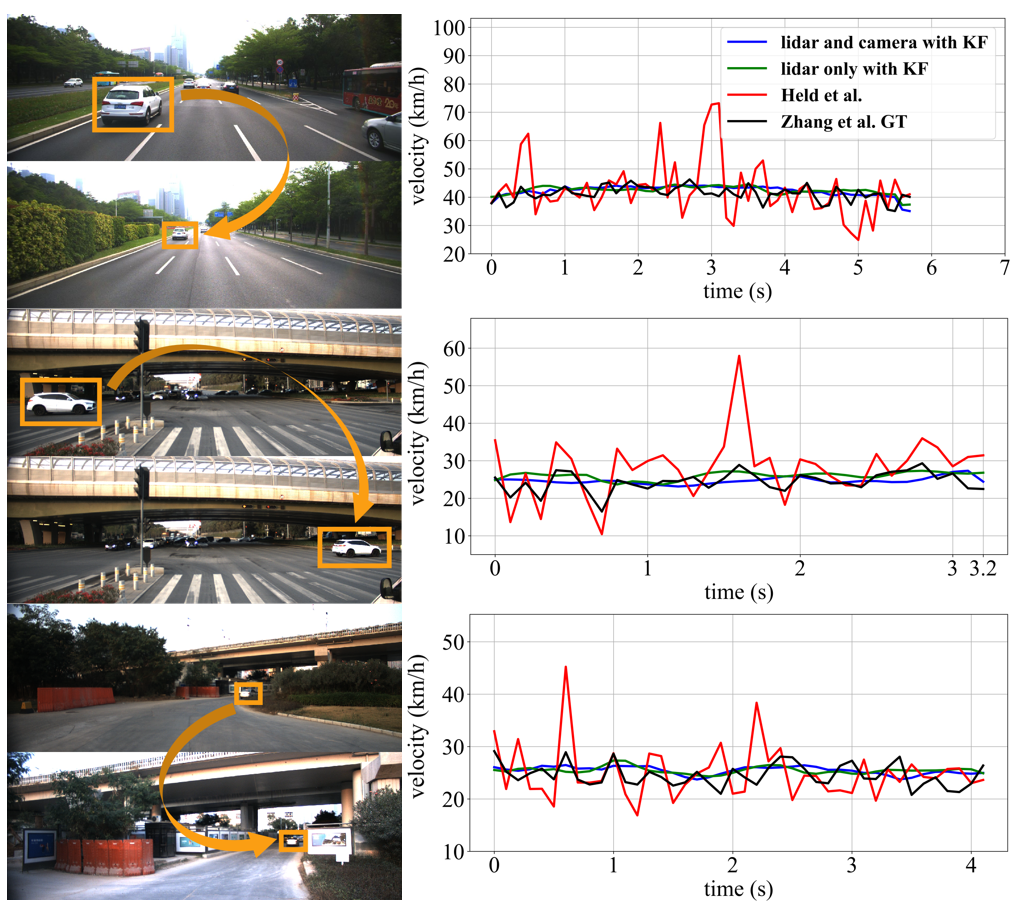}
  \caption{the tracking and velocity estimation performance comparison in different scenarios, from left to right are radial, tangential and turning scenes.}
  \label{fig:detail_tracking_velocity}
\end{figure}

To further validate the estimation accuracy of the proposed approach, ground truth velocities of the tracked vehicles should be compared. Since the direct odometry data from the target cars could not be obtained, we employed the ground truth calculation method proposed by Zhang \textit{et al.} \cite{zhang2020vdo}, where the object bounding box center in global coordinate is computed at every frame and velocities is computed from successive frames. It is worth noting that this method could not provide a precise transient velocity due to the inaccuracy of the bounding box, but the integrated distance over a long distance should be accurate because it simply measures the trace of the moving object in the global coordinate. The error of the bounding box detection would be diluted for a long trace. In Fig. \ref{fig:detail_tracking_velocity} we examined three scenarios: the target car moving in radial direction; the target car moving in tangential direction; and our sensor car is turning. The estimated velocities from different methods are compared. The proposed methods obtained a better match with the ground truth velocity (Zhang \textit{et al.}) than the Held \textit{et al.} method, and maintain the smallest variations among all the methods. Table. \ref{table:our_evaluate_detail} integrates the velocities overtime for distances and the proposed fusion method agrees best with the ground truth, with good crispness score.

\begin{table*}[htbp]
    \centering
  	\begin {tabular}{c p{1.2cm}<{\centering} p{1.2cm}<{\centering} p{1.2cm}<{\centering} p{1.2cm}<{\centering} p{1.2cm}<{\centering} p{1.2cm}<{\centering}}
  		\toprule	
  		-  & \multicolumn{2}{c}{\textbf{\textit{Case Radial}}} & \multicolumn{2}{c}{\textbf{\textit{Case Tangential}}} & \multicolumn{2}{c}{\textbf{\textit{Case Turning}}}   \\ 
  		metrics & \makecell[r]{crisp.} & \makecell[l]{distance($m$)} & \makecell[r]{crisp.} & \makecell[l]{distance($m$)} & \makecell[r]{crisp.} & \makecell[l]{distance($m$)}   \\ 
  		\midrule
  		lidar and camera with KF & \makecell[r]{0.392} & \makecell[l]{68.0}  & \makecell[r]{${0.341}$} & \makecell[l]{22.6} & \makecell[r]{${0.307}$} & \makecell[l]{28.9} \\ 
  		lidar only with KF       & \makecell[r]{${0.395}$} & \makecell[l]{68.3} & \makecell[r]{0.316} & \makecell[l]{23.2} & \makecell[r]{0.288} & \makecell[l]{29.1} \\ 
  		Held \textit{et al.}                     & \makecell[r]{0.203} & \makecell[l]{73.2} & \makecell[r]{0.177} & \makecell[l]{25.9} & \makecell[r]{0.194} & \makecell[l]{29.7} \\ 
  		Zhang \textit{et al.} as GT      & \makecell[r]{-} & \makecell[l]{67.4} & \makecell[r]{-} & \makecell[l]{22.4} & \makecell[r]{-} & \makecell[l]{28.6} \\ 
  		\toprule
  	\end{tabular}
  	
  	\caption{Comparing the various methods with ground truth}
    \label{table:our_evaluate_detail}
\end{table*}

\subsection{Applicability on Rotating Lidars}

\begin{figure}[htbp]
    \flushleft
  \includegraphics[scale=0.35]{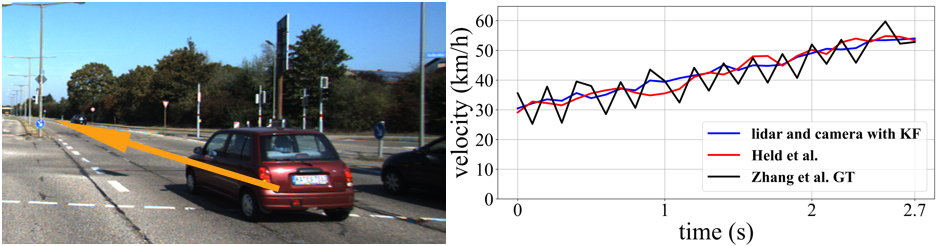}
  \caption{Comparison of performance in KITTI tracking Seq06.}
  \label{fig:kitti_velocity_demo}
\end{figure}

In principle, the proposed method could be applied if we combine a few lidar frames from the rotating lidars and correct the overall distortion. In our implementation, points from the past three frames are merged and corrected for a unified speed. 20 sequences of one object in seq06 of the KITTI object tracking dataset \cite{Geiger2012CVPR} (Velodyne HDL-64 point clouds and synchronized color images) is used and evaluated using the sensor-fusion approach, lidar-only approach and the Held \textit{et al.} approach. The final crispness scores are presented in Table. \ref{table:kitti_evaluation}, which are similar for all three methods because they are essentially correcting the same inter-frame distortions. They also predicted similar velocity estimations as shown in Fig. \ref{fig:kitti_velocity_demo}.

\begin{table}[htbp]
	\caption{crispness score on kitti tracking seq 06}
	\resizebox{0.50\textwidth}{!}
	{
  	\begin {tabular}{c c c c}

  		\toprule	
  		-     & \textbf{\textit{lidar and camera with KF}} &  \textbf{\textit{lidar only with KF}}  & \textbf{\textit{Held \textit{et al.}}} \cite{held2014combining}  \\ 
  		\midrule
  		Crispness Score &  0.340 &  0.342  & 0.336\\ 
  		\toprule
  	\end{tabular}
  	}
\label{table:kitti_evaluation}
\end{table}

\subsection{Runtime Analysis}

Finally a runtime performance analysis is performed. Detection and tracking of multiple moving objects, velocity estimation and point cloud distortion correction were performed simultaneously in successive frames. The Kalman filter takes in the lidar-camera fusion velocity measurements and maintain the best predicted velocity and correct the distortions. The above tasks were evaluated on an Intel i7-10700K CPU running at 3.80 GHz with a breakdown summarized in Table \ref{table:timing}, where a real time performance can be obtained (with less than 6 concurrently tracked objects and frame time is 100 ms).

\begin{table}[htbp]

	\centering \caption{TYPICAL COMPUTING TIME}
	\resizebox{0.5\textwidth}{!}
	{
   	\begin {tabular}{l l}
   		\toprule	
   		Key modules               & Time (ms)            \\ 
   		\midrule
   		Camera detection and optical flow            &  30 *    
   		\\ \cmidrule {1-2}
   		Point cloud optimization                 &   10 **                  \\ 
   		\cmidrule {1-2}
   		KF tracking               &  \textless 1            \\ 
   		\cmidrule {1-2} 
   		Total                     &  41                    \\ 
   		\bottomrule
   		* For all objects in a single frame ($1520\times 568$ resolution) \\
   		** For single object (a car at 20 meters) with 200 points \\

   	\end{tabular}
   	}
\label{table:timing}
\end{table}

\section{CONCLUSIONS}

In this paper we have proposed a robust sensor fusion based velocity estimator and point cloud distortion corrector. By fusing camera and lidar modalities with their unique strength, the distortion that comes from the unique oscillating scan pattern of the emerging lidars can be corrected satisfactorily. With the capability to track moving objects effectively and accurately, we believe this framework is specifically suited for the adoption of the new oscillating lidars on autonomous driving tasks. We hope this open-sourced framework and data would be helpful to both academia and industrial communities that specialize in these areas.

\section*{ACKNOWLEDGMENT}

The authors thank colleagues at Livox Technology for assistance in data collection and discussion.

\printbibliography

\end{document}